\documentclass[runningheads]{llncs}
\usepackage{amsmath}
\usepackage{amssymb}
\usepackage[misc]{ifsym}
\usepackage{graphicx}
\usepackage{color}

\usepackage{algorithm}
\usepackage{algorithmicx}
\usepackage{algpseudocode}
\usepackage{multirow}
\usepackage{array}
\newcolumntype{L}[1]{>{\raggedright\let\newline\\\arraybackslash\hspace{0pt}}m{#1}}
\newcolumntype{C}[1]{>{\centering\let\newline\\\arraybackslash\hspace{0pt}}m{#1}}
\newcolumntype{R}[1]{>{\raggedleft\let\newline\\\arraybackslash\hspace{0pt}}m{#1}}
\begin{document}
\title{Joint Shape Representation and Classification for Detecting PDAC}
\titlerunning{Joint Shape Representation and Classification for Detecting PDAC}
%
%
\author{Fengze Liu\inst{1} \and
Lingxi Xie\inst{1} \and Yingda Xia\inst{1} \and \\
 Elliot Fishman\inst{2} \and
Alan Yuille\inst{1}}
\authorrunning{F. Liu et al.}
%
\institute{ The Johns Hopkins University, Baltimore, MD 21218, USA \and
 The Johns Hopkins University School of Medicine, Baltimore, MD 21287, USA
}
\maketitle              

\begin{abstract}
We aim to detect pancreatic ductal adenocarcinoma (PDAC) in abdominal CT scans, which sheds light on early diagnosis of pancreatic cancer. This is a 3D volume classification task with little training data. We propose a two-stage framework, which first segments the pancreas into a binary mask, then compresses the mask into a shape vector and performs abnormality classification. Shape representation and classification are performed in a {\em joint} manner, both to exploit the knowledge that PDAC often changes the {\bf shape} of the pancreas and to prevent over-fitting. Experiments are performed on $300$ normal scans and $136$ PDAC cases. We achieve a specificity of $90.2\%$ (false alarm occurs on less than $1/10$ normal cases) at a sensitivity of $80.2\%$ (less than $1/5$ PDAC cases are not detected), which show promise for clinical applications.
\end{abstract}

\section{Introduction}
\label{Introduction}

Pancreatic cancer is a major killer causing hundreds of thousands of deaths globally every year. It often starts with a small set of localized cells multiplying themselves out of control and invading other parts of the body. The five-year survival rate of the patient can reach $20\%$~\cite{board2017pancreatic} if the cancer is detected at an early stage, but quickly drops to $5\%$ if it is discovered late and the cancerous cells have spread to other organs~\cite{stewart2017world}. Therefore, early diagnosis of pancreatic cancer can mean the difference between life and death for the patients. 

This paper deals with PDAC, the major type of pancreatic cancer accounting for about $85\%$ of the cases~\cite{stewart2017world}, and attempts to detect it by checking abdominal CT scans. The pancreas, even in a healthy state, is difficult to segment from a CT volume~\cite{roth2015deeporgan}, partly because its 3D shape is irregular~\cite{zhang2017personalized}. The segmentation, particularly for the cancer lesion area, becomes even more challenging when the pancreas is abnormal, {\em e.g.}, cystic~\cite{zhou2017deep}. In recent years, with the development of deep learning frameworks~\cite{krizhevsky2012imagenet}, researchers were able to construct effective deep encoder-decoder networks~\cite{long2015fully} for organ segmentation~\cite{ronneberger2015u} or shape representation~\cite{brock2016generative}, boosting the accuracy of conventional models for a wide range of medical imaging analysis tasks.

The goal of this paper is to discriminate abnormal pancreases from normal ones\footnote{Throughout this paper, an {\em abnormal} pancreas is defined as one suffering from PDAC.}. This is a classification task, but directly training a volumetric classifier may suffer from over-fitting due to limited training data. Inspired by the fact that PDAC often changes the pancreas shape, we set shape representation as an intermediate goal, so as to constrain the learning space and regularize the model. Our framework contains two stages. First, we train an encoder-decoder network~\cite{yu2017recurrent} for voxel-wise pancreas segmentation from CT scans\footnote{To make our approach generalized, we do not assume the tumors are annotated in the training set, and so we do not perform tumor segmentation.}. Second, we use a joint shape representation and classification network to predicts if the patient suffers from PDAC. The weights of the shape representation module are initialized using an auto-encoder~\cite{brock2016generative}\cite{hinton2006reducing}, and then jointly optimized with the classifier. Joint optimization improves classification accuracy at the testing stage.

The radiologists in our team collected and annotated a dataset with $436$ CT scans, including $300$ normal cases and $136$ PDAC cases. Our approach achieves a sensitivity of $80.2\%$ at a specificity of $90.2\%$, {\em i.e.}, finding $4/5$ of abnormal cases with false alarms on only $1/10$ of the normal cases. Some detected PDAC cases contain tiny tumors, which are easily missed by segmentation algorithms and even some professional radiologists. According to the radiologists, our approach can provide auxiliary cues for clinical purposes.

\vspace{-0.2cm}

\section{Detecting PDAC in Abdominal CT Scans}
\label{Approach}

\subsection{The Overall Framework}
\label{Approach:Framework}

A CT-scanned image, $\mathbf{X}$, is a $W\times H\times L$ matrix, where $W$, $H$ and $D$ are the width, height and length of the cube, respectively. Each element in the cube indicates the Hounsfield unit (HU) at the specified position. Each volume is annotated with a binary pancreas mask $\mathbf{S}^\star$ which shares the same dimensionality with $\mathbf{X}$. Our goal is to design a discriminative function ${p\!\left(\mathbf{X}\right)}\in{\left\{0,1\right\}}$, with $1$ indicating that this person suffers PDAC and $0$ otherwise.

Our idea is to decompose the function into two stages. The first stage is a segmentation model $\mathbf{f}\!\left(\cdot\right)$ for voxel-wise pancreas segmentation, {\em i.e.}, where ${\mathbf{S}}={\mathbf{f}\!\left(\mathbf{X}\right)}$. The second stage is a mask classifier $c\!\left(\cdot\right)$ which assigns a binary label to the mask $\mathbf{S}$. To make use of shape information, $c\!\left(\cdot\right)$ is further decomposed into a shape encoder $\mathbf{g}\!\left(\cdot\right)$ which produces a compact vector ${\mathbf{v}}={\mathbf{g}\!\left(\mathbf{S}\right)}$ to depict the shape properties of the binary mask $\mathbf{S}$, and a shape classifier $h\!\left(\cdot\right)$ which determines if the shape vector $\mathbf{v}$ corresponds to a pancreas suffering from PDAC.

\begin{figure}[!t]
\begin{center}
    \includegraphics[width=9.5cm]{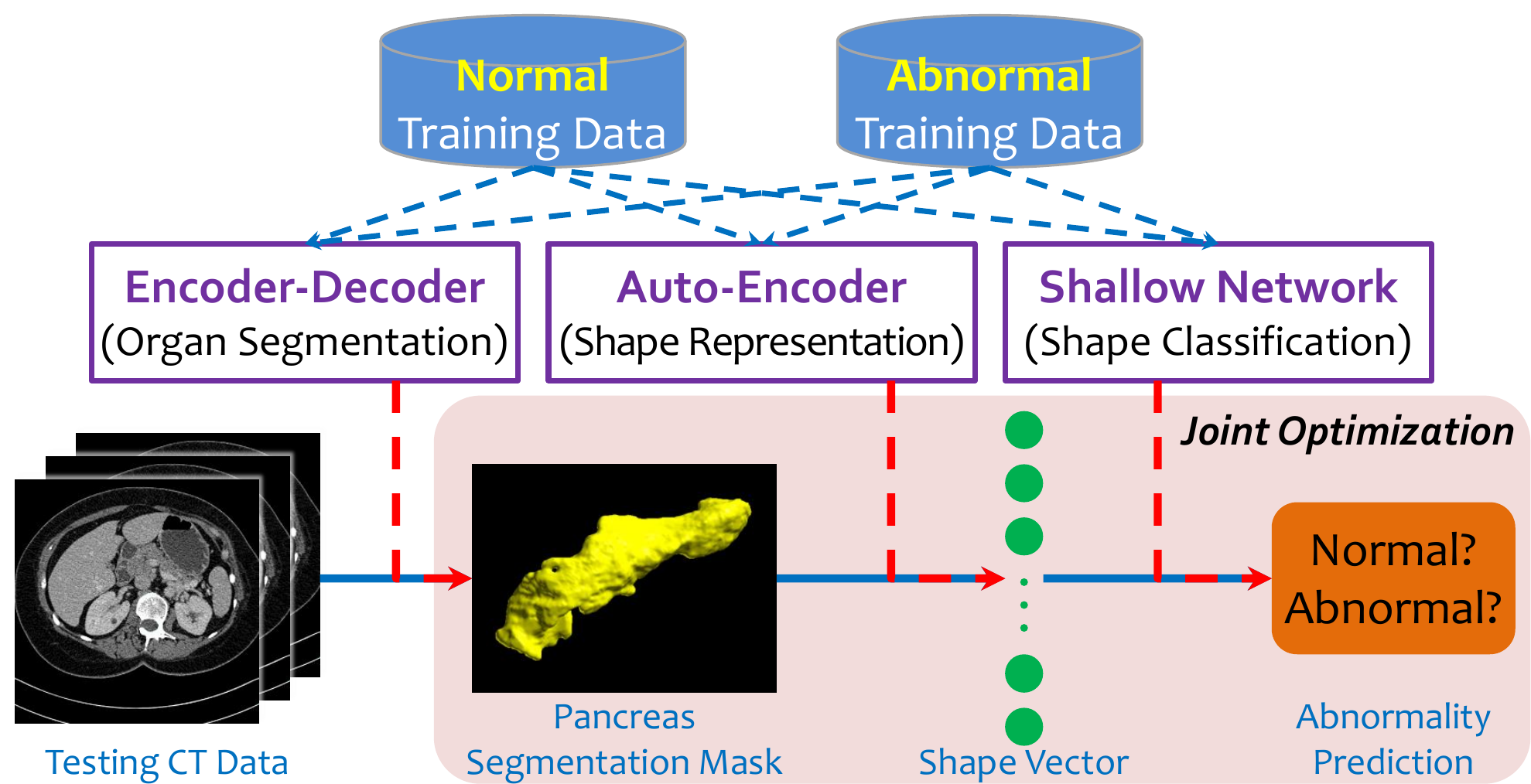}
\end{center}
\caption{
    The overall framework of our approach (best viewed in color).
}
\vspace{-0.2cm}
\label{Fig:Framework}
\end{figure}

Therefore, the overall framework, shown in Figure~\ref{Fig:Framework}, can be written as:
\begin{equation}
\label{Eqn:Framework}
{p\!\left(\mathbf{X}\right)}={c\circ\mathbf{f}\!\left(\mathbf{X}\right)}={h\circ\mathbf{g}\circ\mathbf{f}\!\left(\mathbf{X}\right)}.
\end{equation}
We can of course design an alternative function, {\em e.g.}, a 3D classifier which works on CT image data directly, but our stage-wise model makes use of the prior knowledge from the radiologists, {\em i.e.}, PDAC often changes the shape of the pancreas. This sets up an intermediate goal of optimization and shrinks the search space of our model, which is especially helpful in preventing over-fitting given limited training data. In addition, this also enables us to interpret our prediction. We will show in experiments that, without such prior knowledge, the classifier produces unstable results and less satisfying prediction accuracy.

\vspace{-0.2cm}

\subsection{Pancreas Segmentation by Encoder-Decoder Networks}
\label{Approach:Segmentation}

Our approach starts with an encoder-decoder network for pancreas segmentation. There are typically two choices, which differ from each other in the way of processing volumetric data. The first one applies 2D segmentation networks~\cite{ronneberger2015u}\cite{roth2015deeporgan} from orthogonal planes, while the other one trains a 3D network directly~\cite{milletari2016v} in a patch-based manner. Either method requires cutting volumetric data into 2D slices or 3D patches at both training and testing stages. As a result, the segmentation function ${\mathbf{S}}={\mathbf{f}\!\left(\mathbf{X}\right)}$ cannot be optimized together with the subsequent modules, namely shape representation and classification.

In practice, we apply a recent 2D segmentation approach named RSTN (Recurrent Saliency Transformation Network)~\cite{yu2017recurrent} for pancreas segmentation. It trains three models from the {\em coronal}, {\em sagittal} and {\em axial} planes, respectively. In our own dataset, RSTN works very well, providing an average DSC (Dice Similarity Coefficient) of over $87\%$ for normal pancreas segmentation, and over $70\%$ for abnormal pancreas segmentation. We make two comments here. First, the segmentation accuracy of $87\%$ almost reaches the agreement between two individual annotations by different radiologists. Second, the abnormal pancreases are often more difficult to segment, as their appearance and geometry properties can be changed by PDAC. However, as shown later, such imperfections in segmentation only cause little accuracy drop in abnormality classification.

\subsection{Joint Shape Representation and Classification}
\label{Approach:Classification}

\begin{figure}[!t]
\begin{center}
    \includegraphics[width=12cm]{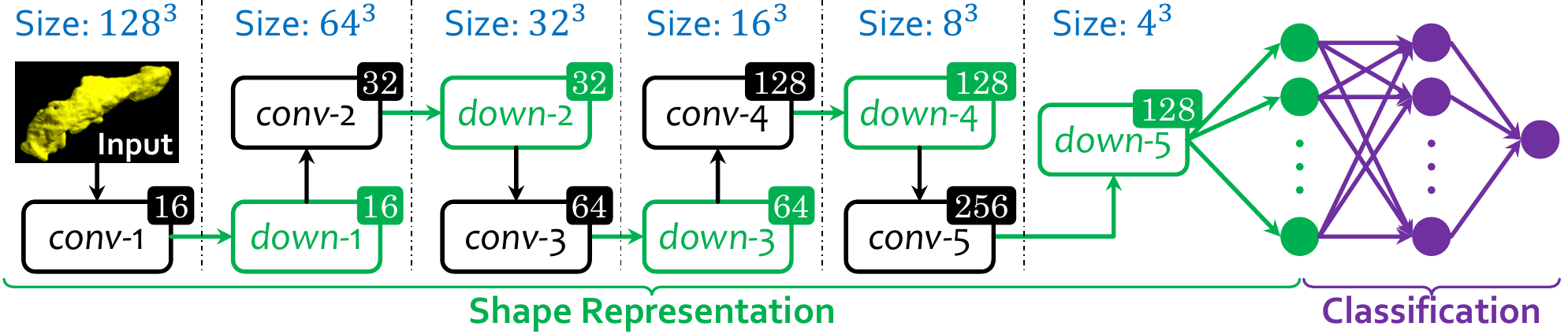}
\end{center}
\vspace{-0.5cm}
\caption{
    Shape representation and classification network (best viewed in color). Each rectangle is a layer, with the number at the upper-right corner indicating the number of channels. Each convolution ({\em conv}) layer contains a set of $3\times3\times3$ kernels, and each down-sampling ({\em down}) layer uses $2\times2\times2$ convolution with a stride of $2$. Batch normalization and ReLU activation are used after all these layers. The last layer in shape representation (the green neurons) is the low-dimensional shape vector, followed by a $2$-layer fully-connected network for classification.
}
\label{Fig:ShapeRepresentation}
\end{figure}

Based on pancreas segmentation ${\mathbf{S}}={\mathbf{f}\!\left(\mathbf{X}\right)}$, it remains to determine the abnormality of this pancreas. We achieve this by first compressing the segmentation mask into a low-dimensional vector ${\mathbf{v}}={\mathbf{g}\!\left(\mathbf{S}\right)}$ to compress $\mathbf{v}$, and then applying a classifier $h\!\left(\cdot\right)$ on top of $\mathbf{v}$.

The shape representation network $\mathbf{g}\!\left(\cdot\right)$ involves down-sampling the segmentation mask gradually. Following~\cite{brock2016generative}, this is implemented by a series of 3D convolutional layers. The detailed network configuration is shown in Figure~\ref{Fig:ShapeRepresentation}. Regarding the dimensionality of the shape vectors ({\em i.e.}, the number of output neurons), a high-dimensional representation carries more information, but also risks over-fitting under limited training data. We analyze this parameter in experiments. Essentially, both segmentation and shape representation networks perform image down-sampling. The former starts with the raw input image and thus requires complicated and expensive computations. The latter, however, is much simpler, with the network much shallower, which processes the entire volume at once. This makes it possible to be optimized together with the classifier.

In the final step, we implement $h\!\left(\cdot\right)$ as a $2$-layer fully-connected network. The simplicity of $h\!\left(\cdot\right)$ aligns with our motivation, {\em i.e.}, the vector $\mathbf{v}$ carries discriminative shape information which is easy to classify. Being a differentiable module, it can the optimized with the shape representation network in a joint manner (details are elaborated below), which brings consistent accuracy gain.

The training process starts by sampling a segmentation mask $\mathbf{S}$ from training data. We first perform slight rotation ($0^\circ$ or $\pm10^\circ$ along three axes individually, $27$ possibilities) as data augmentation, and rescale the region within the minimal bounding box into $128\times128\times128$. Note that direct optimization on $h\circ\mathbf{g}\!\left(\cdot\right)$ cannot guarantee that $\mathbf{g}\!\left(\cdot\right)$ learns shape information. In addition, direct optimization can lead to over-fitting with limited training data, even after data augmentation (see experiments). Hence, we use a two-step method for gradual optimization.

In the first step, we deal with $\mathbf{g}\!\left(\cdot\right)$ by concatenating this module with a decoder network $\tilde{\mathbf{g}}\!\left(\cdot\right)$, which performs reverse operations (all convolutions are replaced by deconvolutions) to restore the compressed vector into the original image. This framework, named an auto-encoder~\cite{brock2016generative}\cite{hinton2006reducing}, can be trained in a weakly-supervised manner, {\em i.e.}, given an input mask $\mathbf{S}$, we can minimize the difference between $\mathbf{S}$ and ${\tilde{\mathbf{S}}}={\tilde{\mathbf{g}}\circ\mathbf{g}\!\left(\mathbf{S}\right)}$ by minimizing the loss function $\mathcal{L}_\mathrm{S}\!\left(\mathbf{S},\tilde{\mathbf{S}}\right)$. This forces the compressed vector $\mathbf{v}$ to store sufficient information in order to restore ${\mathbf{S}}={\tilde{\mathbf{g}}\!\left(\cdot\right)}$. Auto-encoder provides a reasonable initialization for $\mathbf{g}\!\left(\cdot\right)$ in the next step (joint optimization). We use a mini-batch size of $1$ and train the auto-encoder for $40\rm{,}000$ iterations with a fix learning rate of $10^{-6}$.

The second step optimizes $\mathbf{g}\!\left(\cdot\right)$ and $h\!\left(\cdot\right)$ jointly. We use the cross-entropy loss ${\mathcal{L}_\mathrm{C}\!\left(y,p\right)}={y\ln p+\eta\cdot\left(1-y\right)\ln\!\left(1-p\right)}$ where $y$ is the ground-truth and ${p}={h\circ\mathbf{g}\!\left(\mathbf{S}\right)}$ is the predicted confidence. $\eta$ performs class-balancing to avoid model bias. The mini-batch size is still set to be $1$, and we perform a total of $40\rm{,}000$ iterations. We start with a learning rate of $0.0005$, and divide it by $10$ after $20\rm{,}000$ and $30\rm{,}000$ iterations. To maximally preserve stability, we freeze all weights of $\mathbf{g}\!\left(\cdot\right)$ in the first $5\rm{,}000$ iterations, so that the $2$-layer network $h\!\left(\cdot\right)$, initialized as scratch, is reasonably trained before being optimized together with $\mathbf{g}\!\left(\cdot\right)$.

Last but not least, there is an alternative way of jointly optimizing $\mathbf{g}\!\left(\cdot\right)$ and $h\!\left(\cdot\right)$, {\em i.e.}, applying a discriminative auto-encoder~\cite{rolfe2013discriminative}, which preserves the shape restoration loss in the second step and optimizes $\mathcal{L}_\mathrm{S}\!\left(\mathbf{S},\tilde{\mathbf{S}}\right)+\lambda\cdot\mathcal{L}_\mathrm{C}\!\left(y,p\right)$. We do not use this strategy because our ultimate goal is classification -- shape representation is an important cue, but we do not hope the constraints in shape restoration harms classification accuracy. In experiments, we find that a discriminative auto-encoder produces less stable classification accuracy.

\section{Experiments}
\label{Experiments}

\subsection{Dataset and Settings}
\label{Experiments:Dataset}

To the best of our knowledge, there are no publicly available datasets for PDAC diagnosis. We collect a dataset with the help of the radiologists in our team. There are $300$ normal CT scans and $136$ biopsy-proven abnormal (PDAC) cases, and all of them were scanned by the same machine. The pancreas annotation was done by four expert in abdominal anatomy and each case was checked by a experienced board certified Abdominal Radiologist. The spatial resolution of our data is relatively high, {\em i.e.}, the physical distance between the neighboring voxels is $0.5\mathrm{mm}$ in the long axis, and varies from $0.5\mathrm{mm}$ to $1.0\mathrm{mm}$ in the other two axes. We do not use data scanned from other types of machines ({\em e.g.}, the NIH dataset~\cite{roth2015deeporgan}) to avoid dataset bias, {\em i.e.}, the classifier works by simply checking the spatial resolution or other meta-information of the scan.

We use $100$ normal cases for training the RSTN~\cite{yu2017recurrent} and auto-encoder~\cite{brock2016generative} for pancreas segmentation and shape representation, respectively. The remaining $200$ normal and $136$ abnormal scans are first segmented using the RSTN then compressed by the auto-encoder. These examples are randomly split into $4$ folds, each of which has $50$ normal and $34$ abnormal cases. We perform cross-validation, {\em i.e.}, training a classifier on three folds and testing it on the remaining one. We report the sensitivity and specificity of different models.

\subsection{Quantitative Results}
\label{Experiments:Results}

\newcommand{\colwidthA}{1.5cm}
\begin{table}[!btp]
\centering
\begin{tabular}{|l||R{\colwidthA}|R{\colwidthA}||R{\colwidthA}|R{\colwidthA}||R{\colwidthA}|R{\colwidthA}|}
\hline
\multirow{2}{*}{Dimension}
             &                       \multicolumn{2}{c||}{SVM} &                   \multicolumn{2}{c||}{2LN (I)} &                    \multicolumn{2}{c|}{2LN (J)} \\
\cline{2-7}
{}           &                  Sens. &                  Spec. &                  Sens. &                  Spec. &                  Sens. &                  Spec. \\
\hline\hline
$ 128$       &          $73.4\pm 3.1$ &          $87.8\pm 2.9$ &          $77.5\pm 2.2$ &          $87.6\pm 1.5$ &          $79.3\pm 1.0$ &          $89.9\pm 1.0$ \\
\hline
$ 256$       &          $75.0\pm 1.9$ &          $87.6\pm 3.2$ &          $78.2\pm 1.6$ &          $89.1\pm 1.2$ &          $79.0\pm 0.4$ &          $90.5\pm 0.8$ \\
\hline
$ 512$       &          $78.1\pm 1.9$ &          $89.5\pm 1.0$ &          $80.7\pm 1.5$ &          $88.3\pm 1.0$ &          $79.0\pm 0.8$ & $\mathbf{90.9}\pm 0.9$ \\
\hline
$1\rm{,}024$ &          $75.0\pm 0.0$ &          $89.0\pm 0.0$ &          $78.8\pm 0.7$ &          $90.5\pm 0.6$ &          $\mathbf{80.2}\pm 0.5$ & $90.2\pm 0.2$ \\
\hline
\end{tabular}
\caption{
    The sensitivity (sens., $\%$) and specificity (spec., $\%$) reported by different approaches and dimensionalities of shape. We denote the models optimized individually and jointly by (I) and (J), respecitively. All these numbers are the average over $5$ individual runs. 2LN (J) with $1\rm{,}024$-dimensional vectors has the best average performance.
}
\vspace{-0.6cm}
\label{Tab:Results}
\end{table}

Results are summarized in Table~\ref{Tab:Results}. To compare with the joint training strategy, we provide two other competitors, namely a support vector machine (SVM) and the individually-optimized $2$-layer network (equivalent to freezing the parameters in the auto-encoder throughout the entire training process). We observe consistent accuracy gains brought by the proposed approach over both competitors, in particular the $2$-layer network optimized individually. This stresses the importance and effectiveness of joint optimization. Regarding other options, we find that the classification accuracy of our approach either drops or becomes unstable if we (i) train the entire network from scratch; (ii) preserve the shape restoration loss with classification loss; or (iii) do not freeze the weights of the auto-encoder in the early training sections.

In clinics, an important issue to consider is the tradeoff between sensitivity and specificity. A higher sensitivity implies that more abnormal cases are detected, but also brings the price of a lower specificity. Our approach, by simply tuning the classification threshold, can satisfy different requirements. The ROC curves of different models are shown in Figure~\ref{Fig:Visualization}. Using our best model ($1\rm{,}024$-dimensional shape vector with joint optimization), we can achieve a sensitivity of $95\%$ at a specificity of $53.8\%$, or a specificity of $95\%$ at a sensitivity of $67.9\%$.

\subsection{Qualitative Analysis}
\label{Experiments:Analysis}

We first investigate the relationship between pancreas segmentation quality and classification accuracy. Trained on a standalone set of $100$ normal cases, RSTN reports average DSCs of $86.66\%$ and $71.45\%$ on the $200$ testing normal and $136$ abnormal cases, respectively. The radiologists randomly checked around $20$ cases, and verified that our segmentation results, especially on the normal pancreases, have achieved the level of being used for diagnosis. We also use the ground-truth segmentation masks of these $200+136$ pancreases in classification. With $1\rm{,}024$-dimensional shape vectors, the sensitivity and specificity of the SVM classifier are improved by by $9.0\%$ and $2.0\%$, and these numbers for the $2$-layer network are $5.6\%$ and $0.6\%$, respectively. This indicates that the imperfection of abnormal pancreas segmentation mainly causes drops in sensitivity. But, built on top of automatic segmentation, our framework can be applied to a wide range of scenarios where the manual annotation is not available.

Next, we consider the accuracy of shape representation, or more specifically, the similarity between the restored segmentation mask and the original one. It is obvious that a higher dimension in shape representation stores richer information and thus produces more accurate restoration. However, as shown in Table~\ref{Tab:Results}, we do not observe significant gain brought by high dimensionalities. This verifies our assumption, {\em i.e.}, the classifier does not require accurate shape reconstruction. This also explains the advantage of joint optimization, in which the classifier can capture discriminative information from shape representation, and the shape model can also adjust itself to help classification.

\begin{figure}[!t]
\begin{center}
    \includegraphics[width=8.0cm]{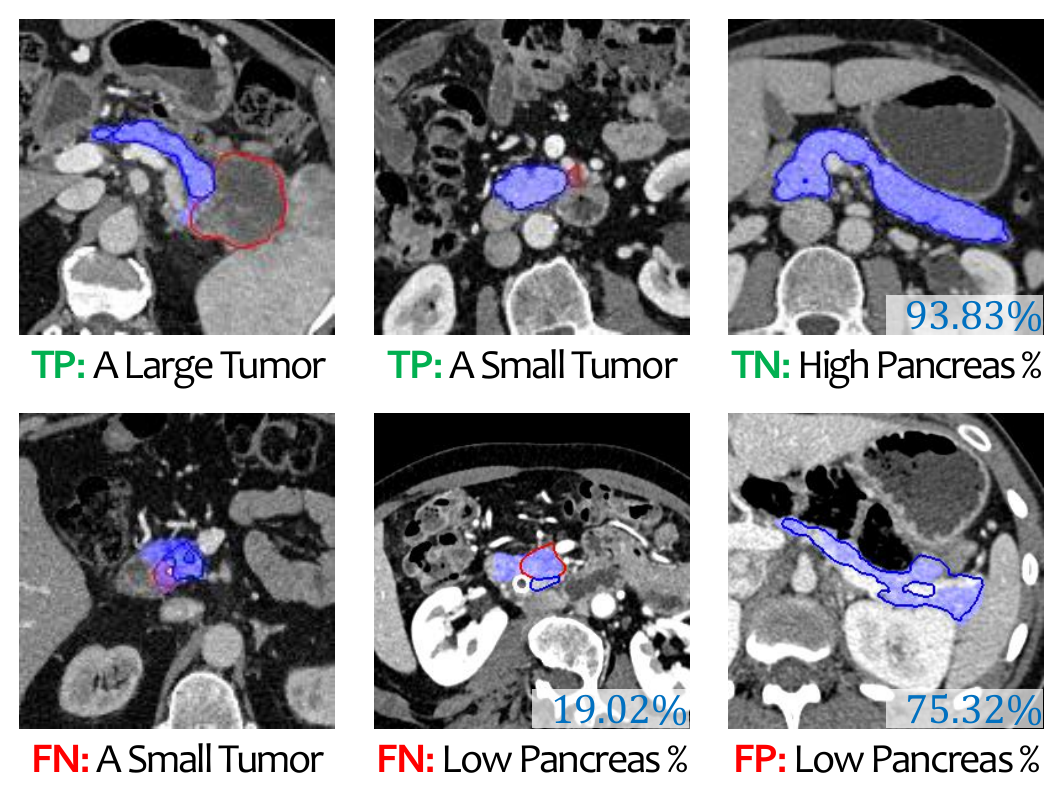}
    \includegraphics[width=3.45cm]{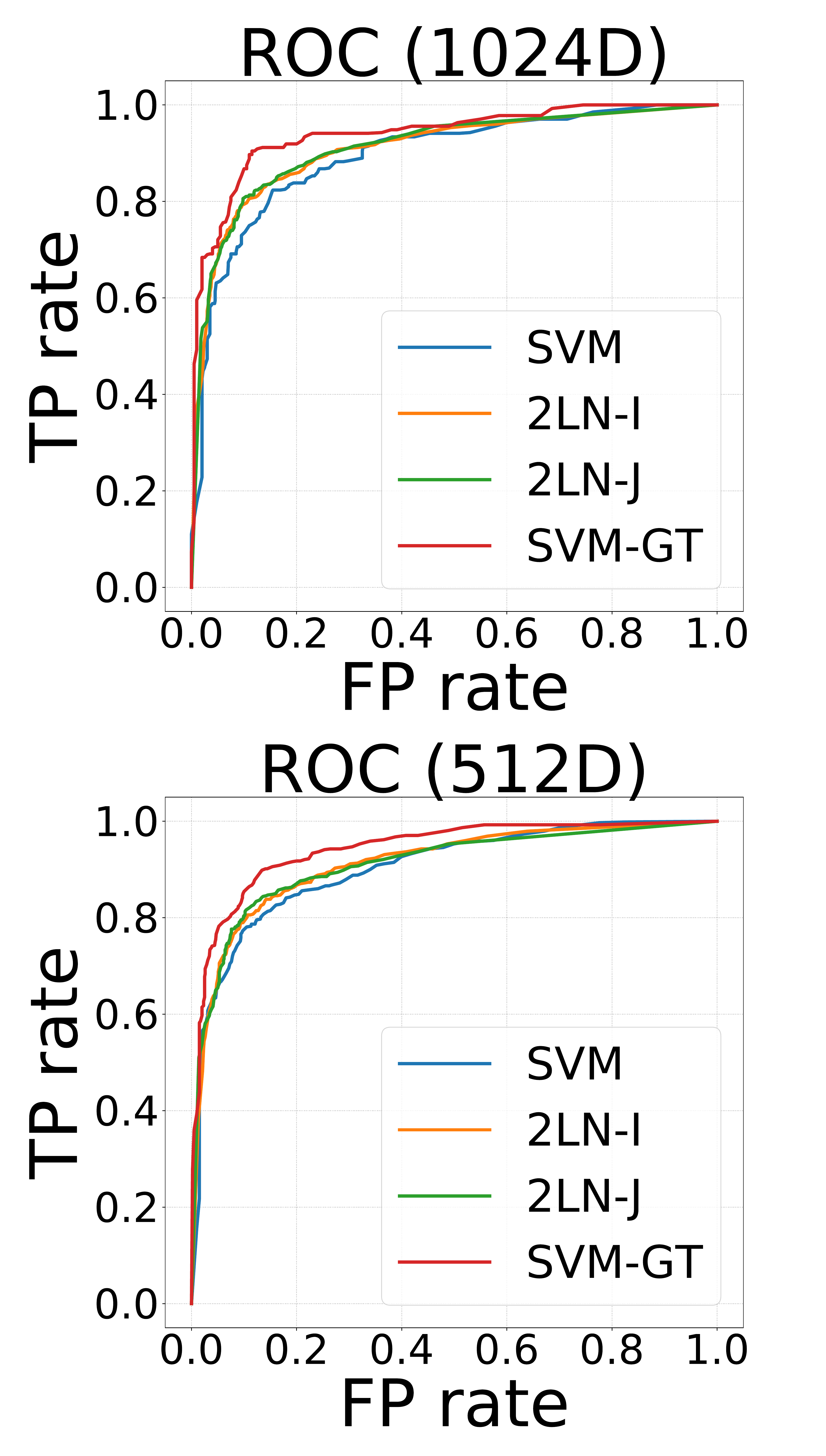}
\end{center}
\vspace{-0.5cm}
\caption{
    Left: classification results by our approach. Right: the ROC curves. Red and blue contours mark the labeled pancreas and tumor, and blue regions indicate the predicted pancreas. TP, TN, FP, FN: for \{true,false\}-\{positive,negative\}, respectively.
}
\label{Fig:Visualization}
\end{figure}

We visualize several successful and failure examples in Figure~\ref{Fig:Visualization}. Our approach is able to detect some cases with tiny tumors which are easily missed even by the radiologists\footnote{The early diagnosis of PDAC is difficult and can be uncertain from CT scans. In our case, the radiologists proved these PDAC cases with biopsy checks. They can easily miss some of these cases if they were not told their abnormality beforehand.}. On the other hand, our approach is likely to fail when the pancreas segmentation is less accurate, leading to a strange pancreas shape which is not seen in training data and thus confuses the classifier. One false-negative and one false-positive cases are shown in Figure~\ref{Fig:Visualization}.

Finally, we point out that there is an alternative to our approach, which directly trains segmentation/detection networks to find the tumors in these PDAC cases. In comparison, our approach has two advantages. First, we do not require the tumors to be annotated in the training data, which is an extremely challenging task. Second, our approach can detect some PDAC cases with very small tumors (which largely changed the shape of the pancreas) that are missed by segmentation. We train a tumor segmentation network individually, and find that more than half of the false negative can be recovered by our approach. This suggests that shape representation serves as an auxiliary cue. However, a clear drawback of our approach is not being able to find the exact position of the lesion area. In all, our approach provides an important cue (shape), and it can be integrated with other cues in the future towards more accurate diagnosis, {\em e.g.}, when voxel-wise tumor annotations are available, we can incorporate pancreas/tumor segmentation into our joint optimization framework.

\section{Conclusions}
\label{Conclusions}

Our approach is motivated by knowledge from surgical morphology, which claims that the PDAC can be discovered by observing the shape change of the pancreas. We first use an encoder-decoder network to obtain pancreas segmentation, and design a joint framework for shape representation and classification. We initialize shape representation using an auto-encoder, and optimize it with the classifier in a joint manner.

In experiments, our approach achieved a sensitivity of $80.2\%$ with a specificity of $90.2\%$. It even detected several challenging cases which are easily missed by the radiologists. Given a larger amount of training data, we can expect even higher performance. Our future research directions also involve adding other cues ({\em e.g.}, tumor segmentation) and training the entire framework in a joint manner.

 \bibliographystyle{splncs04}
 \bibliography{ref.bib}
\end{document}